\documentclass{llncs}
\usepackage{graphicx}
\usepackage{amsmath,amssymb} 
\usepackage{color}
\usepackage[width=122mm,left=12mm,paperwidth=146mm,height=193mm,top=12mm,paperheight=217mm]{geometry}

\usepackage{soul}
\usepackage{hyperref}

\usepackage{floatrow}
\usepackage{subfiles}

\newfloatcommand{capbtabbox}{table}[][\FBwidth]

\usepackage{xspace}
\usepackage{color}
\usepackage{booktabs}
\usepackage{subcaption}
\usepackage{microtype}
\usepackage{enumitem}
\setitemize[0]{leftmargin=10pt}

\newcommand{\eg}{e.g.\@}
\newcommand{\etal}{et al.\@}
\newcommand{\etc}{etc.\xspace}
\newcommand{\ie}{i.e.\@}

\newlength\tikzfigwidth
\newlength\tikzfigheight

\def\realspace{\mathbb{R}} 

\newcommand{\mat}[1]{\ensuremath{\mathbf{#1}}} 

\newcommand{\chis}[2][non]{\ensuremath{\chi^2 \ifthenelse{\equal{#1}{non}}{}{ \left(#1,#2\right)}}} 
\newcommand{\Gammaf}[1][non]{\ensuremath{\Gamma\ifthenelse{\equal{#1}{non}}{}{ \left( #1 \right)}}} 

\newcommand{\degree}[1][non]{\ensuremath{\ifthenelse{\equal{#1}{non}}{^\circ}{#1^\circ}}} 

\def\image{I}
\def\imagePix{\image}
\def\imh{h}
\def\imw{w}

\def\meanRGB{m}

\def\semsegFg{S^{\textrm{fg}}}
\def\semsegFgPix{\semsegFg}

\def\mask{M}
\def\maskPix{\mask}
\def\imageMasked{\image^{\mask}}
\def\imageMaskedPix{\imageMasked}
\def\maskCls{\mask^{\textrm{cls}}}

\def\numCls{C}
\def\numFgCls{\numCls^{\textrm{fg}}}
\def\numBgCls{\numCls^{\textrm{bg}}}

\def\semsegBg{S^{\textrm{bg}}}
\def\depthmapBg{D^{\textrm{bg}}}

\def\bevmapInit{B^{\textrm{init}}}
\def\bevh{k}
\def\bevw{l}
\def\bevmapFinal{B^{\textrm{final}}}
\def\bevmapSim{B^{\textrm{sim}}}
\def\bevmapMask{\mat{M}}
\def\bevmapOSM{B^{\textrm{osm}}}
\def\bevmapOSMWarped{\hat{B}^{\textrm{osm}}}
\def\batchsize{m}

\def\warpParams{\theta}

\def\camIntrinsic{\mat{K}}

\def\lossSym{\mathcal{L}} 
\newcommand{\discr}[1]{\ensuremath{d\left(#1\right)}} 
\newcommand{\warp}[2]{\ensuremath{w\left(#1;#2\right)}} 

\def\paramDiscr{\Theta_{\textrm{discr}}}

\makeatletter
\renewcommand{\paragraph}{%
  \@startsection{paragraph}{4}%
  {\z@}{0.5ex \@plus 1ex \@minus .2ex}{-1em}%
  {\normalfont\normalsize\bfseries}%
}
\makeatother

\newcommand\blfootnote[1]{%
  \begingroup
  \renewcommand\thefootnote{}\footnote{#1}%
  \addtocounter{footnote}{-1}%
  \endgroup
}

\begin{document}
\pagestyle{headings}
\mainmatter

\title{Learning to Look around Objects for Top-View Representations of Outdoor Scenes}

\titlerunning{Learning to Look around Objects for Top-View Representations of Road Scenes}
\authorrunning{Samuel Schulter, Menghua Zhai, Nathan Jacobs, Manmohan Chandraker}
\author{Samuel Schulter$^{1,\dagger}$ $\quad$ Menghua Zhai$^{2,\dagger}$ $\quad$ Nathan Jacobs$^2$\\ Manmohan Chandraker$^{1,3}$}
\institute{NEC-Labs$^1$, Computer Science University of Kentucky$^2$, UC San Diego$^3$}

\maketitle

\begin{abstract}
  Given a single RGB image of a complex outdoor road scene in the perspective view, we address the novel problem of estimating an occlusion-reasoned semantic scene layout in the top-view.  This challenging problem not only requires an accurate understanding of both the 3D geometry and the semantics of the visible scene, but also of occluded areas.  We propose a convolutional neural network that learns to predict occluded portions of the scene layout by looking around foreground objects like cars or pedestrians.  But instead of hallucinating RGB values, we show that directly predicting the semantics and depths in the occluded areas enables a better transformation into the top-view.  We further show that this initial top-view representation can be significantly enhanced by learning priors and rules about typical road layouts from simulated or, if available, map data.  Crucially, training our model does not require costly or subjective human annotations for occluded areas or the top-view, but rather uses readily available annotations for standard semantic segmentation.  We extensively evaluate and analyze our approach on the KITTI and Cityscapes data sets.
\end{abstract}


\section{Introduction}
\label{sec:inroduction}
Visual completion is a crucial ability for an intelligent agent to navigate and interact with the three-dimensional (3D) world.  Several tasks such as driving in urban scenes, or a robot grasping objects on a cluttered desk, require innate reasoning about unseen regions.  A top-view or bird's eye view (BEV) representation\footnote{We use the terms ``top-view'' and ``bird's eye view'' interchangeably.} of the scene where occlusion relationships have been resolved is useful in such situations~\cite{sam:Gupta17a}.  It is a compact description of agents and scene elements with semantically and geometrically consistent relationships, which is intuitive for human visualization and precise for autonomous decisions.

\blfootnote{$\dagger$ indicates equal contribution}

\begin{figure}[t]
\begin{center}
  \includegraphics[width=1.0\linewidth]{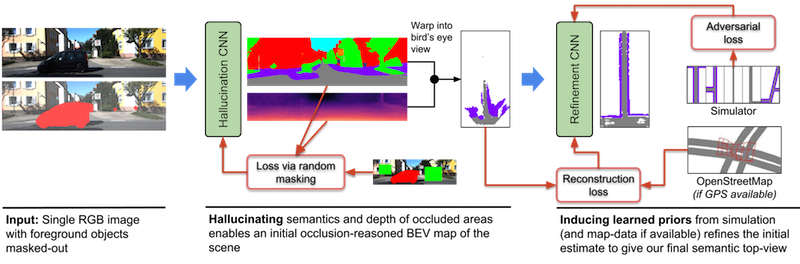}
\end{center}
\vspace{-0.7cm}
\caption{Given a single RGB image of a typical street scene (left), our approach creates an \textbf{occlusion-reasoned semantic map of the scene layout in the bird's eye view}.  We present a CNN that can hallucinate depth and semantics in areas occluded by foreground objects (marked in red and obtained via standard semantic segmentation), which gives an initial but noisy and incomplete estimate of the scene layout (middle). To fill in unobserved areas in the top-view, we further propose a refinement-CNN that induces learning strong priors from simulated and OpenStreetMap data (right), which comes at no additional annotation costs.}
\label{fig:teaser}
\end{figure}

In this work, we derive such top-view representations through a novel framework that simultaneously reasons about geometry and semantics from just \emph{a single RGB image}, which we illustrate in the particularly challenging scenario of outdoor road scenes.
The focus of this work lies in the estimation of the scene layout, although foreground objects can be placed on top using existing 3D localization methods~\cite{sam:Song14a,sam:Zia15a}.
Our learning-based approach estimates a geometrically and semantically consistent spatial layout even in regions hidden behind foreground objects, like cars or pedestrians, without requiring human annotation for occluded pixels or the top-view itself.
Note that human supervision for such occlusion-reasoned top-view maps is likely to be subjective and of course, expensive to procure.  Instead, we derive supervisory signals from readily available annotations for semantic segmentation in the perspective view, a depth sensor or stereo (for visible areas) and a knowledge corpus of typical road scenes via simulations and OpenStreetMap data.  Figure~\ref{fig:teaser} provides an illustration.

Specifically, in Section~\ref{sec:inpaint_seg_depth}, we propose a novel CNN that takes as input an image with occluded regions (corresponding to foreground objects) masked out, and estimates the segmentation labels and depth values over the entire image, essentially \emph{hallucinating distances and semantics in the occluded regions}.  In contrast to standard image in-painting approaches, we operate in the semantic and depth spaces rather than the RGB image space.  Section~\ref{sec:inpaint_seg_depth} shows how to train this CNN without additional human annotations for occluded regions.  The hallucinated depth map is then used to map the hallucinated semantic segmentation of each pixel into the bird's eye view, see Section~\ref{sec:map_to_3d}.

This initial prediction can be incomplete and erroneous, for instance, since BEV pixels far away from the camera can be unobserved due to limited image resolution or due to imperfect depth estimation.  Thus, Section~\ref{sec:refine_bev} proposes a refinement and completion neural network to leverage easily obtained training data from \emph{simulations that encode general priors and rules} about road scene layouts.  Since there is no correspondence between actual images and simulated data, we employ an {\em adversarial loss} for teaching our CNN a generative aspect about typical layouts.  When GPS is available for training images, we also demonstrate how \emph{map data provides an additional training signal} for our models.  We demonstrate this using OpenStreetMap (OSM)~\cite{OpenStreetMap}.  Maps provide rough correspondence with RGB images through the GPS location, but it can be noisy and lacks information on scene scale, besides mislabels in the map itself.  We handle these issues by {\em learning a warping function} that aligns OSM data with image evidence using a variant of spatial transformer network~\cite{sam:Jaderberg15a}.  Note that a single RGB image is used at test time, with simulations or OSM limited to training.

In Section~\ref{sec:experiments}, we evaluate our proposed semantic BEV synthesis on the KITTI~\cite{sam:Geiger13a} and Cityscapes~\cite{sam:Cordts16a} datasets.  For a quantitative evaluation, we manually annotate validation images with the scene layout in both the perspective and the top-view, which is a time-consuming and error-prone process but again highlights the benefit of our method that resorts only to readily available annotations.  Since, to the best of our knowledge, no prior work exists solving this problem in a similar setup to allow a fair comparison, we comprehensively evaluate with several baselines to study the role of each module.  Our experiments consider roads and sidewalks for layout estimation, with cars and persons as occluding foreground objects, although extensions to other semantic classes are straightforward in future work.  While not our focus, we visualize a simple application in Section \ref{sec:exp_foreground_objects} to include foreground objects such as cars and pedestrians in our representation.  We observe qualitatively meaningful top-view estimates, which also obtain low errors on our annotated test set (which will be released upon publication).


\section{Related Work}
\label{sec:related_work}
General scene understanding is one of the fundamental goals of computer vision and many approaches exist that tackle this problem from different directions.

\paragraph{Indoor:} Recent works like~\cite{sam:Armeni16a,sam:Liu15a,sam:Song17a} have shown great progress by leveraging strong priors about and and large-scale data sets of indoor environments.  While indoor scene understanding can rely on strong assumptions like a Manhattan world layout, our work focuses on less constrained outdoor driving scenarios.

\paragraph{Outdoor:}
Scene understanding for outdoor scenes has received a lot of interest in recent years~\cite{sam:Dhiman16a,sam:Guo12a,sam:Sturgess09a,sam:Wojek13a,sam:Zia13a}, especially due to applications like driver assistance systems or autonomous driving.
Wang~\etal~\cite{sam:Wang15a} propose a conditional random field that infers 3D object locations, semantic segmentation as well as a depth reconstruction of the scene from a single geo-tagged image, which also enables the use of OSM data.  At test time, their approach requires as input accurate GPS and map information.  In contrast, we require only the RGB image at test time.
Seff and Xiao~\cite{sam:Seff16a} leverage OpenStreetMap (OSM) data to predict several road layout attributes from a single image, like the distance to an intersection, drivable directions, heading angle, \etc
While we also leverage OSM for training our models and make predictions only from a single RGB image, we infer a full semantic map in the top view instead of a discrete set of attributes.

\paragraph{Top-view representations:}
Sengupta~\etal~\cite{sam:Sengupta12a} derive a top-view representation by relating semantic segmentation in perspective images with a ground plane with a homography.  However, this is a simplifying (flat-world) assumption where non-flat objects will produce artifacts in the ground plane, like shadows or cones.
To alleviate these artifacts, they aggregate semantics over multiple frames.  However, removing all artifacts would require viewing objects from many different angles.  In contrast, our approach enables reasoning about occlusion from just a single image, which is enabled by automatically learned and context-dependent priors about the world.
Geiger~\etal~\cite{sam:Geiger14a} represent road scenes with a complex model in the bird's eye view.  However, input to the model comes from multiple sources (vehicle tracklets, vanishing points, scene flow, \etc) and inference requires MCMC, while our approach efficiently computes the BEV representation from just a single image.  Moreover, their hand-crafted parametric model might not account for all possible scene layouts, whereas our approach is non-parametric and thus more flexible.
M\'attyus~\etal~\cite{sam:Mattyus16a} combine perspective and top-view images to estimate road layouts and Zhai~\etal~\cite{sam:Zhai17a} predict the semantic layouts of top-view images by learning the transformation between the perspective and the top-view.
Gupta~\etal~\cite{sam:Gupta17a} demonstrate the suitability of a BEV representation for mapping and planning, even though it is not explicitly learned.

\paragraph{Occlusion reasoning:}
Most recent works in this area focus on occlusions of foreground objects and employ complex hand-crafted models~\cite{sam:Dhiman16a,sam:Wojek13a,sam:Zia13a,sam:Zia15a}.  In contrast, we estimate the layout of a scene occluded by foreground objects.  Guo and Hoiem~\cite{sam:Guo12a} employ a scene parsing approach that retrieves existing shapes from training data based on visible pixels.  Our approach learns to hallucinate occluded areas and does not rely on an existing and fixed set of polygons from training data.  Liu~\etal~\cite{sam:Liu16a} also hallucinate the semantics and depth of regions occluded by foreground objects.  However, (i) their approach relies on a hand-crafted graphical model while ours is learning-based and (ii) they assume sparse depth from a laser scanner as input, while we estimate depth from a single RGB image (the sparse depth maps are actually ground truth for training our models).


\section{Generating bird's eye view representations}
\label{sec:generating_bird_eye_view_representations}
We now present our approach for transforming a single RGB image in the perspective view into an occlusion-reasoned semantic representation in the bird's eye view, see Figure~\ref{fig:teaser}.  We take as input an image $\image \in \realspace^{\imh \times \imw \times 3}$ with spatial dimension $\imh$ and $\imw$ and a semantic segmentation $\semsegFg \in \realspace^{\imh \times \imw \times \numCls}$ of the \emph{visible} scene, where $\numCls$ is the number of categories.  Note that any semantic segmentation method can be used and we rely on the recently proposed pyramid scene parsing (PSP) network~\cite{sam:Zhao17a}.  $\semsegFg$ provides the location of foreground objects that occlude the scene.  In this work, we consider foreground objects like cars or pedestrians as occluders but other definitions are possible as well.

To reason about these occlusions, we define a masked image $\imageMasked$, where pixels of foreground objects have been removed.  In Section~\ref{sec:inpaint_seg_depth}, we propose a CNN that takes $\imageMasked$ as input and hallucinates the depth as well as the semantics of the entire image, including occluded pixels.
The occlusion-reasoned depth map $\depthmapBg$ allows us to map the occlusion-reasoned semantic segmentation $\semsegBg$ into 3D and then into the bird's eye view (BEV), see Section~\ref{sec:map_to_3d}.

While this initial BEV map $\bevmapInit$ is already better than mapping the non-occlusion reasoned semantic map $\semsegFg$ into 3D, there can still be unobserved or erroneous pixels.  In Section~\ref{sec:refine_bev}, we thus propose a CNN that learns priors from simulated data to further improve our representation.  If a GPS signal is available, OpenStreetMap (OSM) data can be additionally included as supervisory signal.

\subsection{Learning to see around foreground objects}
\label{sec:inpaint_seg_depth}
An important step towards an occlusion-reasoned representation of the scene is to infer the semantics and the geometry behind foreground objects.

\paragraph{Masking:}
Given the semantic segmentation $\semsegFg$, we define the mask of foreground pixels as $\mask \in \realspace^{\imh \times \imw}$, where a pixel in the mask $\maskPix_{ij}$ is $1$ if and only if the segmentation at that pixel $\semsegFgPix_{ij}$ belongs to any of the foreground classes.
Otherwise, the pixel in the mask is $0$.
In order to inform the CNN about which pixels have to be in-painted, we apply the mask on the input RGB image and define each pixel in the masked input $\imageMasked$ as
\begin{equation*}
  \imageMaskedPix_{ij} =
  \begin{cases}
    \meanRGB,       & \text{if } \maskPix_{ij} = 1 \\
    \imagePix_{ij},  & \text{otherwise,}
  \end{cases}
\end{equation*}
where $\meanRGB$ is the mean RGB value of the color range, such that after normalization the input to the CNN is zero for those pixels.
Given $\imageMasked$, we extract a feature representation by applying ResNet-50~\cite{sam:He16a}.  Similar to recent semantic segmentation literature~\cite{sam:Zhao17a}, we replace strided convolutions with dilated convolutions~\cite{sam:Yu16a} to increase the feature representation from $\frac{\imh}{32} \times \frac{\imw}{32}$ to $\frac{\imh}{8} \times \frac{\imw}{8}$.

\begin{figure}[t]
\begin{center}
  \begin{subfigure}[t]{0.68\textwidth}
    \includegraphics[width=0.9\textwidth]{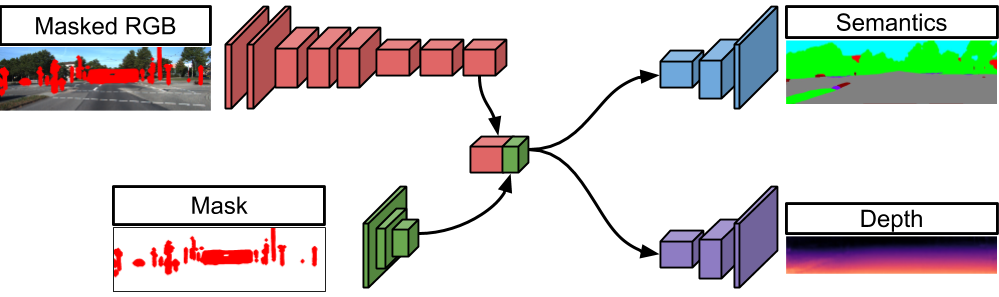}
    \caption{}
    \label{fig:inpaint_details_a}
  \end{subfigure}
  \begin{subfigure}[t]{0.30\textwidth}
    \centering
    \includegraphics[width=1.0\textwidth]{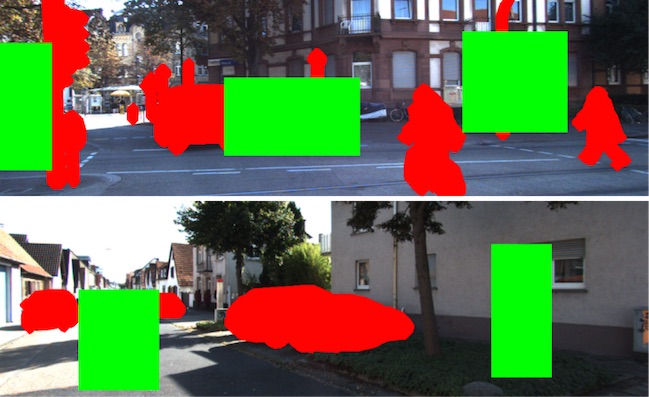}
    \caption{}
    \label{fig:inpaint_details_b}
  \end{subfigure}
\end{center}
\vspace{-0.6cm}
\caption{
  \textbf{(a)} The \emph{inpainting CNN} first encodes a masked image and the mask itself.  The extracted features are concatenated and two decoders predict semantics and depth for visible and occluded pixels.
  \textbf{(b)} To train the inpainting CNN we ignore foreground objects as no ground truth is available (red) but we \emph{artificially add masks (green)} over background regions where full annotation is already available.
}
\label{fig:inpaint_details}
\end{figure}

In addition to masking the input image, we explicitly provide the mask as input to the CNN for two reasons:
(i) While the value $m$ becomes 0 after centering the input of the CNN, other visible pixels might still share the same value and confuse the training of the CNN.
(ii) An explicit mask input allows encoding more information like the category of the occluded pixel.
We thus define another mask $\maskCls \in \realspace^{\imh \times \imw \times \numFgCls}$, where $\numFgCls$ is the number of foreground classes and each channel corresponds to one of them.
We encode $\maskCls$ with a small CNN and fuse the resulting feature with the one from the masked image, see Figure~\ref{fig:inpaint_details_a}.

\paragraph{Hallucination:}
We then put two decoders on the fused feature representation of $\imageMasked$ and $\maskCls$ for predicting semantic segmentation and the depthmap of the occlusion-free scene.
For semantic segmentation, we again use the PSP module~\cite{sam:Zhao17a}, which is particularly useful for in-painting where contextual information is crucial.
For depth prediction, we follow~\cite{sam:Laina16a} in defining the network architecture.
Both decoders are followed by a bilinear upsampling layer to provide the output at the same resolution as the input, see Figure~\ref{fig:inpaint_details_a}.
While traditional in-painting methods fill missing pixels with RGB values, note that we directly go from an RGB image to the in-painted semantics and the geometry of the scene, which has two benefits:
(1) The computational costs are smaller as we avoid the (in our case) unnecessary detour in the RGB space.
(2) The task of in-painting in the RGB space is presumably harder than in-painting semantics and depth as there is no need for predicting any texture or color.

\paragraph{Training:}
We train the proposed CNN in a supervised way. However, as mentioned before, it would be very costly to annotate the semantics and particularly the geometry behind foreground objects.  We thus resort to an alternative that only requires standard semantic segmentation and depth ground truth.  Because our desired ground truth is unknown for real foreground objects in the masked input image $\imageMasked$, we do not infer any loss at those pixels.  However, we augment $\imageMasked$ with additional randomly sampled masks, but for which we still have ground truth, see Figure~\ref{fig:inpaint_details_b}.  In this way, we can teach our CNN to hallucinate occluded areas of the input image without acquiring costly human annotations.
Note that an alternative to masking regions in the input image is to paste real foreground objects into the scene.  However, this strategy requires separate instances of foreground objects cropped at the semantic boundaries and a good understanding of the scene geometry for generating a realistic looking training image.

\begin{figure}[t]
\floatbox[{\capbeside\thisfloatsetup{capbesideposition={right,top},capbesidewidth=0.48\textwidth}}]{figure}[\FBwidth]
{\caption{The process of mapping the semantic segmentation with corresponding depth first into a 3D point cloud and then into the bird's eye view.  The red and blue circles illustrate corresponding locations in all views.}\label{fig:map2bev_details}}
{\includegraphics[width=0.5\textwidth]{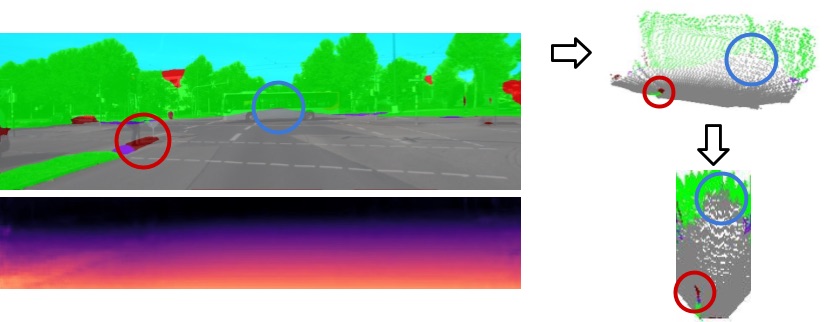}}
\vspace{-0.5cm}
\end{figure}

\vspace{-0.25cm}
\subsection{Mapping into the bird's eye view}
\label{sec:map_to_3d}
Given the depth map $\depthmapBg$ and the intrinsic camera parameters $\camIntrinsic$, we can map each coordinate of the perspective view into the 3D space.
We drop the z-coordinate (height axis) for each 3D point and assign x and y coordinates to the closest integer, which gives us a mapping into bird's eye view representation.
We use this mapping to transfer the class probability distribution of each pixel in the perspective view, \ie, $\semsegBg$, into the bird's eye view, which we denote $\bevmapInit \in \realspace^{\bevh \times \bevw \times \numBgCls}$, where $\numBgCls$ is the number of background classes and $\bevh$ and $\bevw$ are the spatial dimensions.  Throughout the paper, we use $\bevh=128$ and $\bevw=64$ pixels that we relate to $60 \times 30$ meters in the point cloud.  For all points that are mapped to the same pixel in the top view, we average the corresponding class distribution.
Figure~\ref{fig:map2bev_details} illustrates the geometric transformation.

Note that $\bevmapInit$ is our first occlusion-reasoned semantic representation in the bird's eye view.
However, $\bevmapInit$ also has several remaining issues.
Some pixels in $\bevmapInit$ will not be assigned any class probability, especially those far from the camera due to image foreshortening in the perspective view.
Imperfect depth prediction is also an issue because it may assign a well classified pixel in the perspective view a wrong depth value, which puts the point into a wrong location in top-view.  This can lead to unnatural arrangements of semantic classes in $\bevmapInit$.

\subsection{Refinement with a Knowledge Corpus}
\label{sec:refine_bev}
\begin{figure}[t]\centering
  \begin{subfigure}[t]{0.16\textwidth}
    \centering
    \includegraphics[trim={47.5cm 37cm 0 0}, clip, width=0.9\textwidth]{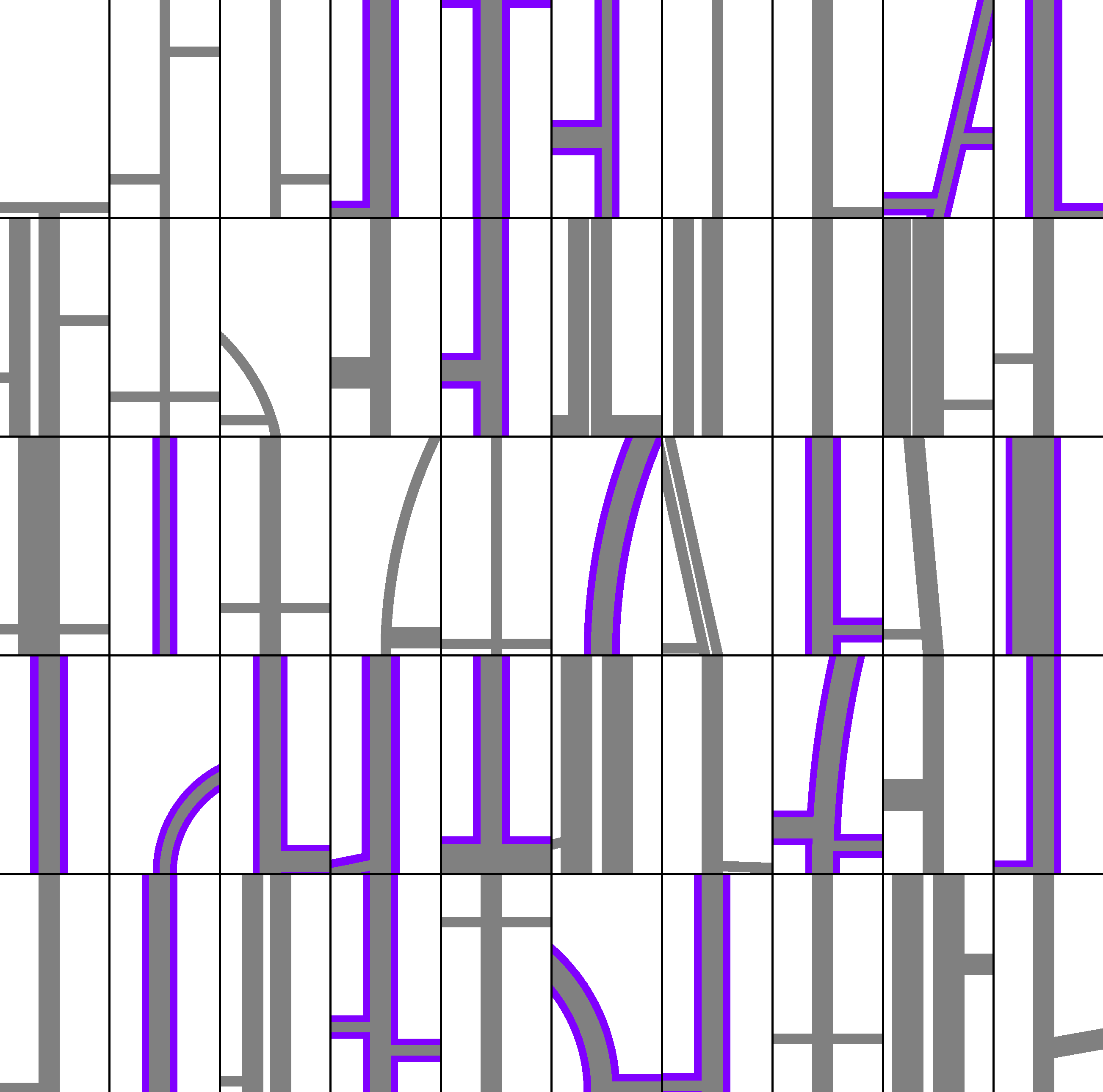}
    \caption{}
    \label{fig:simulator_examples}
  \end{subfigure}
  \begin{subfigure}[t]{0.36\textwidth}
    \centering
    \includegraphics[width=0.9\textwidth]{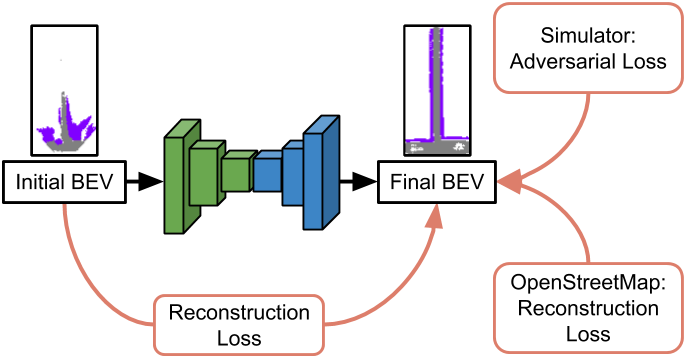}
    \caption{}
    \label{fig:refine_net_a}
  \end{subfigure}
  \begin{subfigure}[t]{0.44\textwidth}
    \centering
    \includegraphics[width=0.9\textwidth]{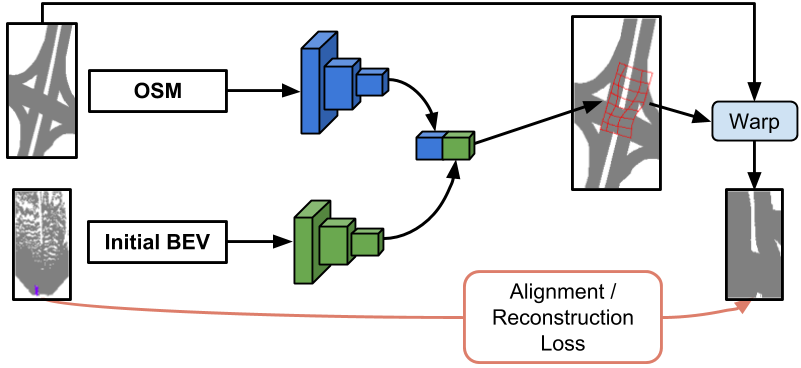}
    \caption{}
    \label{fig:refine_net_b}
  \end{subfigure}
\vspace{-0.2cm}
\caption{
  \textbf{(a) Simulated road shapes} in the top-view.
  \textbf{(b) The refinement-CNN} is an encoder-decoder network receiving three supervisory signals: self-reconstruction with the input, adversarial loss from simulated data, and reconstruction loss with aligned OpenStreetMap (OSM) data.
  \textbf{(c) The alignment CNN} takes as input the initial BEV map and a crop of OSM data (via noisy GPS and yaw estimate given). The CNN predicts a warp for the OSM map and is trained to minimize the reconstruction loss with the initial BEV map.
}
\label{fig:refine_net}
\end{figure}

To remedy the above mentioned issues, we propose a refinement CNN that takes $\bevmapInit$ and predicts the final output $\bevmapFinal \in \realspace^{\bevh \times \bevw \times \numBgCls}$, which has the same dimensions as $\bevmapInit$.  The refinement CNN has an encoder-decoder structure with a fully-connected bottleneck layer, see Figure~\ref{fig:refine_net_a}.
The main difficulty in training the refinement CNN is the lack of semantic ground truth data in the bird's eye view, which is very hard and costly to annotate.  In the following we present two sources of supervisory signals that are easy to acquire.

\paragraph{Simulation:}
The first source of information we leverage is a simulator that renders the semantics of typical road scenes in the bird's eye view.  The simulator models roads with different types of intersections, lanes and sidewalks, see Figure~\ref{fig:simulator_examples} for some examples.  Note that it is easy to create such a simulator as we do not need to model texture, occlusions or any perspective distortions in the scene.  A simple generative model about road topology, number of lanes, radius for curved roads, \etc is enough.
Since there is no correspondence with the real training data, we rely on an adversarial loss~\cite{sam:Arjovsky17a} between predictions of the refinement CNN $\bevmapFinal$ and data from the simulator $\bevmapSim$
\begin{equation*}
  \lossSym^{\textrm{sim}} = \sum_{i=1}^{\batchsize} \discr{\bevmapFinal_{i}; \paramDiscr} - \sum_{i=1}^{\batchsize} \discr{\bevmapSim_{i}; \paramDiscr} \;,
\end{equation*}
where $\batchsize$ is the batch size and $\discr{.; \paramDiscr}$ is the discriminator function with parameters $\paramDiscr$.  Note that $\discr{.; \paramDiscr}$ needs to be a K-Lipschitz function~\cite{sam:Arjovsky17a}, which is enforced by gradient clipping on the parameters $\paramDiscr$ during training.  While any other variant of adversarial loss is possible, we found \cite{sam:Arjovsky17a} to provide the most stable training.  The adversarial loss injects prior information about typical road scene layouts and remedies errors of $\bevmapInit$ like unobserved pixels or unnatural shapes of objects due to depth or semantic prediction errors.

Since $\lossSym^{\textrm{sim}}$ operates without any correspondence, the refinement network needs additional regularization to not deviate too much from the actual input, \ie, $\bevmapInit$.  We add a reconstruction loss between $\bevmapInit$ and $\bevmapFinal$ to define the final loss as $\lossSym = \lossSym^{\textrm{sim}} + \lambda \cdot \lossSym^{\textrm{reconst}}$ with
\begin{equation*}
  \lossSym^{\textrm{reconst}} = \frac{\|(\bevmapInit - \bevmapFinal) \odot \bevmapMask\|^2}{\sum_{ij} \bevmapMask} \;,
\end{equation*}
where $\odot$ is an element-wise multiplication and $\bevmapMask \in \realspace^{\bevh \times \bevw}$ is a mask of 0's for unobserved pixels in $\bevmapInit$ and 1's otherwise.

\begin{figure}[t]\centering
  \vspace{-0.3cm}
  \begin{subfigure}[t]{0.30\textwidth}\centering
    \includegraphics[width=0.8\textwidth]{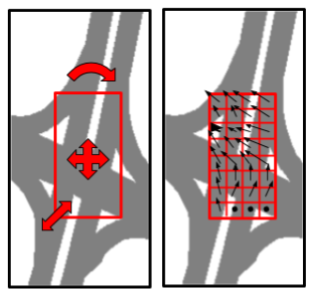}
    \caption{}
    \label{fig:osm_warps_schematic}
  \end{subfigure}
  \begin{subfigure}[t]{0.68\textwidth}\centering
    \includegraphics[width=0.8\textwidth]{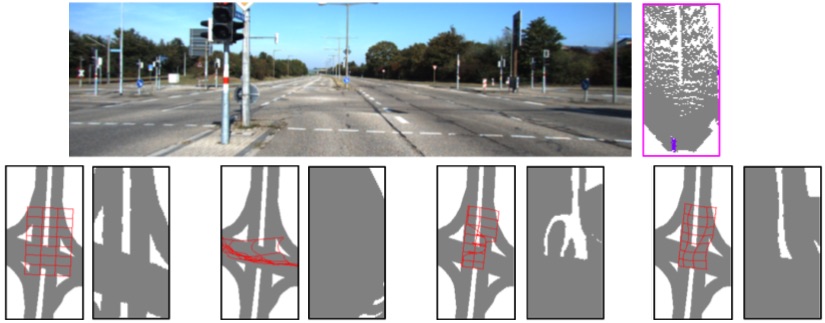}
    \caption{}
    \label{fig:osm_warps_example}
  \end{subfigure}
  \vspace{-0.3cm}
  \caption{
    \textbf{(a)} We use a composition of similarity transform (left, ``box'') and a non-parametric warp (right, ``flow'') to align noisy OSM with image evidence.  \textbf{(b, top)} Input image and the corresponding $\bevmapInit$.  \textbf{(b, bottom)} Resulting warping grid overlaid on the OSM map and the warping result for 4 different warping functions, respectively: ``box'', ``flow'', ``box+flow'', ``box+flow (with regularization)''.  Note the importance of composing the transformations and the induced regularization.
  }
  \label{fig:osm_warping}
\end{figure}

\paragraph{OpenStreetMap data:}
Driving imagery often comes with a GPS signal and an estimate of the driving direction, which enables the use of OpenStreetMap (OSM) data as another source of supervisory signal for the refinement CNN.
The most simple approach is to render the OSM data for the given location and angle, $\bevmapOSM$, and define a reconstruction loss with $\bevmapFinal$ as $\lossSym^{\textrm{OSM}} = \|\bevmapFinal - \bevmapOSM\|^{2}$.  This loss can be included into the final loss $\lossSym$ in addition to or instead of $\lossSym^{\textrm{reconst}}$.  In any case, $\lossSym^{\textrm{OSM}}$ ignores noise in the GPS and the direction estimate as well as imperfect renderings due to annotation noise and missing information in OSM.

We therefore propose to align the initial OSM map $\bevmapOSM$ with the semantics and geometry observed in the actual RGB image with a warping function $\bevmapOSMWarped = \warp{\bevmapOSM}{\warpParams}$ parameterized by $\warpParams$.  We use a composition of a similarity transformation implemented as a parametric spatial transformer (handling translation, rotation, and scale; denoted ``Box'') and a non-parametric warp implemented as bilinear sampling (handling non-linear misalignments due to OSM rendering; denoted ``Flow'')~\cite{sam:Jaderberg15a}, see Figure~\ref{fig:osm_warping}.  We minimize the masked reconstruction between $\bevmapOSMWarped$ and the initial BEV map $\bevmapInit$,
\begin{equation*}
  \warpParams^* = \arg\min_{\warpParams} \frac{\| (\bevmapInit - \warp{\bevmapOSM}{\warpParams}) \odot \bevmapMask \|^2}{\sum_{ij} \bevmapMask} + \lambda_2 \Gamma(\warp{\bevmapOSM}{\warpParams}) + \lambda_3 \|\warpParams\|_2^2\;,
\end{equation*}
where $\warp{.}{\warpParams}$ is differentiable~\cite{sam:Jaderberg15a}, and $\Gamma(.)$ is a low-pass filter similar to~\cite{zhou2017unsupervised,Vijayanarasimhan17}, and $\|.\|^2_2$ the squared $\ell_2$-norm, both acting as regularizing functions.  The hyper-parameters $\lambda_2$ and $\lambda_3$ are manually set.

To minimize the alignment error the first choice is non-linear optimization, \eg, LBFG-S~\cite{byrd1995limited}.  However, we found this to produce satisfactory results only for parts of the data, while a significant portion would require hand-tuning of several hyper-parameters.  This is mostly due to noise in the initial BEV map $\bevmapInit$ as well as the rendering $\bevmapOSM$.  An alternative, which proved to be more stable and easy to realize, is to learn a function that predicts the warping parameters, which has the benefit that the predictive function can implicitly leverage other examples of $(\bevmapInit, \bevmapOSM)$ pairs in the training corpus.  We thus train a CNN that takes $\bevmapInit$ and $\bevmapOSM$ as inputs and predicts the warping parameters $\warpParams$ by minimizing the alignment error.  Also, we can either train this CNN separately or jointly with the refinement CNN, thus providing different training signals for the refinement module.  We evaluate these options in Section~\ref{sec:exp_refine}.  Figure~\ref{fig:refine_net_b} illustrates the process of aligning the OSM data.


\section{Experiments}
\label{sec:experiments}
Our quantitative and qualitative evaluation focuses on occlusion reasoning via hallucination in the perspective view (Section~\ref{sec:exp_inpaint}) and scene completion via the refinement network in the bird's eye view (Section~\ref{sec:exp_refine}).

\paragraph{Datasets:}
Creating the proposed BEV representation requires data for learning the parameters of the modules described above.  Importantly, the only supervisory signal that we need is semantic segmentation (human annotation) and depth (LiDAR or stereo), although not both are required for the same input image.  Both KITTI~\cite{sam:Geiger13a} and Cityscapes~\cite{sam:Cordts16a} fulfill our requirements.  Both data sets come with a GPS signal and a yaw estimate of the driving direction, which allows us to additionally leverage OSM data during training.

The KITTI~\cite{sam:Geiger13a} data set contains many sequences of typical driving scenarios and contains accurate GPS location and driving direction as well as a 3D point cloud from a laser scanner.
However, annotation for semantic segmentation is scarce.
We create two versions of the data set based on segmentation annotation:
\emph{KITTI-Ros} consists of 31 sequences (14201 frames) for training, where 100 of them have semantic annotation, and of 9 sequences (4368) for validation, where 46 images are annotated for segmentation.
The segmentation ground truth comes from~\cite{sam:Ros15a}.
\emph{KITTI-Anon} consists of 31 sequences (16273 frames) for training, where 1074 images have semantic annotations, and 9 sequences (2296 frames) for validation, where 233 images are annotated for segmentation.
We collected this segmentation ground truth on our own and will make it publicly available.

The \emph{Cityscapes} data set~\cite{sam:Cordts16a} contains $2975$ training and $500$ validation images, all of which are fully annotated for semantic segmentation and are provided as stereo image pairs.  For ease of implementation, we rely on a strong stereo method~\cite{sam:Zbontar16a} to serve as our training signal for depth, although unsupervised methods exists for direct training from stereo images~\cite{sam:Garg16a,sam:Godard17a}.  GPS location and heading are also provided, although accuracy is lower compared to KITTI.

\paragraph{Validation data:}
For a quantitative evaluation of occlusion reasoning in the perspective view as well as in the bird's eye view, we manually annotated all validation image of the three data sets that also have semantic segmentation ground truth.  We asked annotators to draw the scene layout by hand for the categories ``road'' and ``sidewalk''.  Other pixels are annotated as ``background''.  We will also make this test data publicly available.

\begin{figure}[t]
\begin{floatrow}
\ffigbox[\FBwidth]{%
  \includegraphics[width=6.5cm]{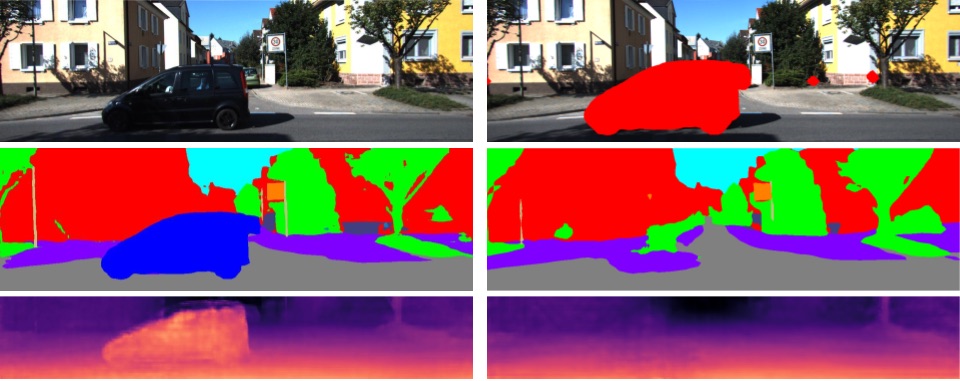}
}{%
  \vspace{-0.2cm}\caption{Qualitative example of our hallucination CNN: Semantics and depth without (left) and with (right) hallucination.}\label{fig:inpaint_results_qualitative}%
}
\capbtabbox[\Xhsize]{%
  \scriptsize
\begin{tabular}{l c c c c}
  \toprule
  Method                  & \multicolumn{2}{c|}{random-boxes} & \multicolumn{2}{c}{human-gt}  \\
                          & \multicolumn{2}{c|}{hidden}       & \multicolumn{1}{c|}{visible} & \multicolumn{1}{c}{hidden}  \\
                          & IoU & \multicolumn{1}{c|}{ARD}    & \multicolumn{1}{c|}{IoU} & IoU \\
  \toprule
  RGB-inpaint             & \textbf{68.83} & .1428  & 79.25 & 55.79 \\
  Direct                  & 64.63 & \textbf{.1413}  & \textbf{81.12} & \textbf{60.06} \\
  \midrule
  RGB-only                & 63.07 & .1440  & 79.71 & 60.77 \\
  + mask                  & 63.47 & \textbf{.1435}  & 80.14 & 60.24 \\
  + cls-encode            & \textbf{64.79} & .1453  & \textbf{80.63} & \textbf{61.06} \\
  \bottomrule
\end{tabular}


}{%
  \vspace{-0.2cm}\caption{Hallucination results for two general in-painting strategies and different mask encodings.}%
  \label{tbl:inpaint_results_alldatasets}
}
\end{floatrow}
\vspace{-0.4cm}
\end{figure}

\paragraph{Implementation details:}
We train our in-painting models with a batch size of 2 for 80k iterations with ADAM~\cite{sam:Kingma15a}.  The initial learning rate is $0.0002$, which is decreased by a factor of 10 for the last 20k iterations.  The refinement network is trained with a batch size of 64 for 80k iterations and a learning rate of $0.0001$.

\subsection{Occlusion Reasoning by Hallucination}
\label{sec:exp_inpaint}
Here we analyze our hallucination CNN proposed in Section~\ref{sec:inpaint_seg_depth}, which targets at in-painting the semantics and depth of areas occluded by foreground objects.  To the best of our knowledge, there is no prior art that can serve as a \emph{fair} comparison point.  Although \cite{sam:Liu16a} addresses the same task, their approach assumes sparse depth information as input, which serves as ground truth in our approach.  Nevertheless, we have created fair baselines that justify our design choices.

\paragraph{Evaluation protocol:}
We split our evaluation protocol into two parts.  First, we follow \cite{sam:Liu16a} by randomly masking out background regions in the input and evaluate the predictions of the hallucination CNNs (random-boxes).  For this case, note that evaluation can be done for all semantic classes and depth.  While this is the only possible evaluation without human annotation for occluded areas, the sampling process may not resemble objects realistically.  Thus, we also evaluate with our newly acquired annotations (human-gt) for the categories ``road'' and ``sidewalk'', which was not done in~\cite{sam:Liu16a}.  We measure mean IoU for segmentation and absolute relative distance (ARD) for depth estimation as in~\cite{sam:Laina16a}.

\paragraph{Semantics \& depth space versus RGB space:}
We compare our hallucination CNN with a baseline that takes the traditional approach of in-painting and operates in the RGB pixel space.  This baseline consists of two CNNs, one for in-painting in the RGB space and one for semantic and depth prediction.  For a fair comparison, we equip both CNNs with the same ResNet-50 feature extractor.  For RGB-space in-painting, we use the same decoder structure as for depth prediction but with 3 output channels and train it with the random mask sampling strategy.  The second CNN has the exact same architecture as our hallucination CNN and is trained without masking inputs but instead uses the already in-painted RGB images.
From Table~\ref{tbl:inpaint_results_alldatasets} we can see that the proposed direct hallucination network outperforms in-painting in the RGB space for depth prediction and segmentation with the human-provided ground truth while it trails for segmentation of all categories with random boxes.  The reason for the inferior performance might be missing context information that is available to the baseline by the RGB-space supervision.  However, note that the proposed architecture is twice as efficient, since in-painting and prediction of semantics and depth are obtained in the same forward pass.  Qualitative examples of our direct hallucination CNN are given in Figure~\ref{fig:inpaint_results_qualitative}.

\paragraph{Mask-encoding:}
We also analyze different variants of how to encode the foreground mask as input to the proposed hallucination CNN.  Table~\ref{tbl:inpaint_results_alldatasets} demonstrates the beneficial impact of explicitly encoding the foreground mask (``+mask'') in addition to masking the RGB image (``RGB-only''), as well as providing the class information of the foreground objects inside the mask (``+cls-encode'').

\begin{table}[t]\centering\scriptsize
  \begin{subtable}{0.5\textwidth}
  \begin{tabular}{l c c c}
  \toprule
  Setting (KITTI-Anon)        & Road    & Sidewalk    & Mean   \\
  \toprule
  BEV-init                    & 58.13   & 29.33       & 43.73  \\
  Refine-heuristic            & 67.93   & 30.12       & 49.02  \\
  Simulation                  & 66.98   & 29.73       & 48.36  \\
  Simulation+OSM              & \textbf{68.89}   & \textbf{30.35}       & \textbf{49.62}  \\
  \midrule
  no halluc.                   & 51.85   & 24.76       & 38.31  \\
  no halluc. (refine)          & 65.67   & 25.91       & 45.79  \\
  no depth pred.               & 44.54   &  8.61       & 26.58  \\
  no depth pred. (refine)      & 46.11   &  7.73       & 26.92  \\
  \bottomrule
\end{tabular}


  \caption{}
  \label{tbl:bev_main_results_kittianon}
  \end{subtable}%
  \begin{subtable}{0.5\textwidth}
  \begin{tabular}{l l c c c}
  \toprule
  Dataset                 & Setting                   & Road    & Sidewalk    & Mean   \\
  \toprule
  KITTI-                  & BEV-init                  & 56.93   & 40.71       & 48.82  \\
  Ros                     & Refine-heuristic          & 69.59   & 41.31       & 55.45  \\
                          & Simulation                & 62.96   & 43.19       & 53.08  \\
                          & Simulation+OSM            & \textbf{71.82}   & \textbf{44.77}       & \textbf{58.29}  \\
      \toprule
  City-                   & BEV-init                  & 51.40   & 17.47       & 34.43  \\
  scapes                  & Refine-heuristic          & 52.06   & 17.22       & 34.64  \\
                          & Simulation                & 52.89   & 17.89       & 35.39  \\
                          & Simulation+OSM            & \textbf{56.46}   & \textbf{19.60}       & \textbf{38.03}  \\
  \bottomrule
\end{tabular}


  \caption{}
  \end{subtable}
  \vspace{-0.3cm}
  \caption{\textbf{(a)} Results on the KITTI-Anon data set showing the impact of the refinement module with simulated and OSM data compared to $\bevmapInit$ and a simple refinement heuristic.  We also show the impact of hallucination and depth prediction.  \textbf{(b)} Results for KITTI-Ros and Cityscapes.}
  \label{tbl:bev_main_results_alldatasets}
\end{table}

\begin{table}[t]
\floatbox[{\capbeside\thisfloatsetup{capbesideposition={right,top},capbesidewidth=0.48\textwidth}}]{table}[\FBwidth]
{\caption{An ablation study of the proposed BEV-refinement module.  We analyze different types of warping functions and OSM alignment optimization strategies.}\label{tbl:bev_ablation_results_kittiinternal}}
{\scriptsize\begin{tabular}{l l c c c}
  \toprule
  Experiment     & Setting                   & Road    & Sidewalk    & Mean   \\
  \toprule
  Warping-       & Box                       & 64.77   & 30.51       & 47.64  \\
  method         & Flow                      & 66.03   & \textbf{30.74}       & 48.39  \\
                 & Box+Flow                  & \textbf{68.89}   & 30.35       & \textbf{49.62}  \\
  \midrule
  Warp-          & LBFGS                     & 22.31   & 29.24       & 25.78  \\
  optimization   & CNN                       & 63.91   & 29.19       & 46.55  \\
                 & CNN-joint                 & \textbf{68.89}   & \textbf{30.35}       & \textbf{49.62}  \\
  \bottomrule
\end{tabular}

}
\end{table}

\begin{figure}[t]\centering
\includegraphics[width=1.0\textwidth]{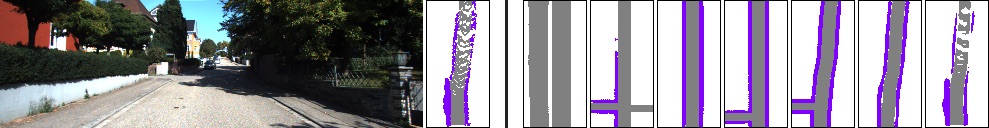}
\vspace{-0.6cm}
\caption{One example of the influence of $\lambda$ (trade-off between adversarial and reconstruction loss, \ie, OSM and simulation data).  (left) input image and $\bevmapInit$;  (right) 6 final BEV maps for $\lambda = \{0, 1, 5, 100, 500, 1000, 10^6\}$.  One can see that higher $\lambda$ leads to higher alignment with $\bevmapInit$.}
\label{fig:bev_lambdasweep_examples}
\end{figure}

\subsection{Refining the BEV representation}
\label{sec:exp_refine}
We now evaluate the refinement model described in Section~\ref{sec:refine_bev} on all three data sets with the acquired annotations in the bird's eye view.  The evaluation metric again is mean IOU for the categories ``road'' and ``sidewalk''.
We compare four models:
(1) The initial BEV map, without refinement.
(2) A refinement heuristic, where missing semantic information at pixel $(i,j)$ is filled with the semantics of the closest pixels in y-direction towards the camera.
(3) The proposed refinement module with simulated data and the self-reconstruction loss.
(4) The refinement module with the additional OSM-reconstruction loss.
Table~\ref{tbl:bev_main_results_alldatasets} clearly shows that the combination of simulated and aligned OSM data provides the best supervisory signal for the refinement module on all three data sets.  Interestingly, the refinement heuristic is a strong competitor but this is probably because evaluation is limited to only ``road'' and ``sidewalk'', where simple rules are often correct.  This heuristic will likely fail for classes like ``vegetation'' and ``building''.  Importantly, all refinement strategies improve upon the initial BEV map.

\begin{figure}[t]\centering\vspace{-0.1cm} \includegraphics[width=0.95\textwidth]{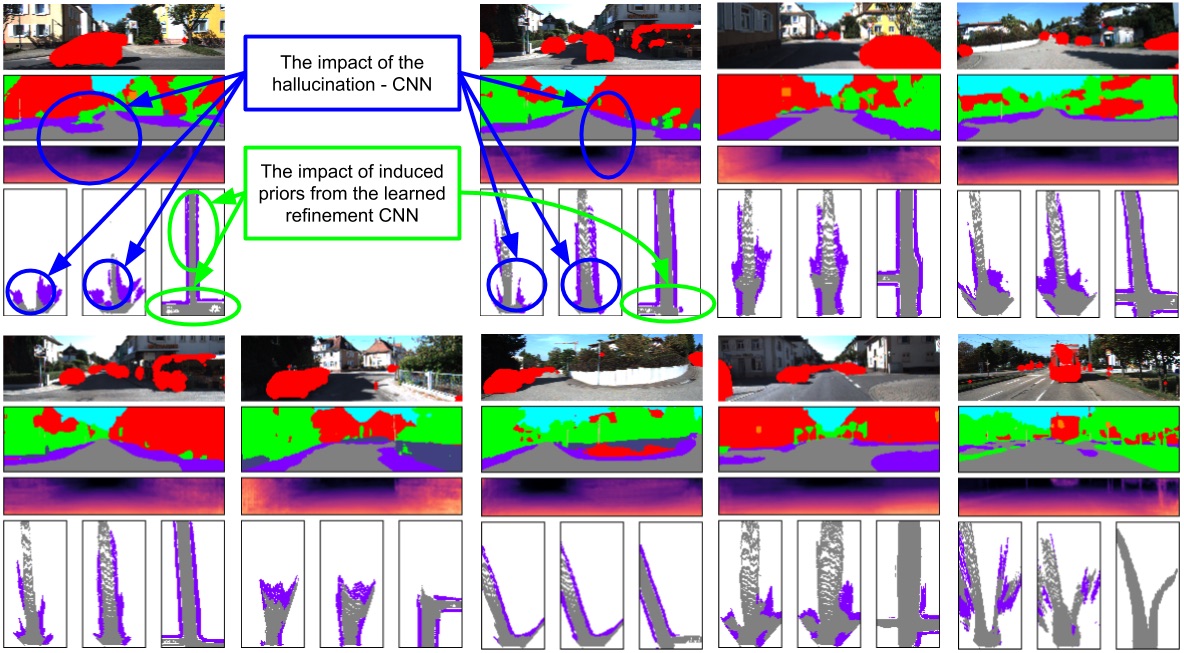}
  \vspace{-0.25cm}
  \caption{\textbf{Examples of our BEV representation}.  Each one shows the masked RGB input, the hallucinated semantics and depth, as well as three BEV maps, which are (from left to right), The BEV map without hallucination, with hallucination, and after refinement.  The last example depicts a failure case.}
\label{fig:bev_qualitative_examples}%
\end{figure}

Because no fair comparison point to prior art is available to us, we further analyze two alternative baselines on the KITTI-Anon data set.

\paragraph{Importance of hallucination:}
We train a refinement module that takes as input BEV maps that omit the hallucination step (``No halluc.'').  To create this BEV map, we train a joint segmentation and depth prediction network (same architecture as for hallucination) with standard foreground annotation and map the semantics of background classes into the BEV map as described in Section~\ref{sec:map_to_3d}.
Table~\ref{tbl:bev_main_results_alldatasets} shows that avoiding the hallucination step hurts the performance.  Note that the proposed refinement CNN recovers most errors for roads, while the relative performance drop for sidewalks is larger.  We believe this is due to long stretches of non-occluded roads in the KITTI data set.  Sidewalks, on the other hand, are typically more occluded due to parked cars and pedestrians.

\paragraph{Importance of depth prediction:}
We train a CNN that takes as input the RGB image in the perspective view and directly predicts the BEV map, without depth prediction (``No depth pred.'').  The CNN extracts basic features with ResNet-50~\cite{sam:He16a}, applies strided convolutions for further down-sampling, a fully-connected layer resembling a transformation from 2D to 3D, and transposed convolutions for up-sampling into the BEV dimensions.  To create a training signal for this network, we map ground truth segmentation with the ground truth depth data (LiDAR) into the bird's eye view.  On top of the output of this CNN, we still apply the proposed refinement module for a fair comparison.
The importance of depth prediction becomes clearly evident from Table~\ref{tbl:bev_main_results_alldatasets}.  In this case, not even the refinement-CNN is able to recover.  While there can be better architectures for directly predicting a semantic BEV map from the perspective view than our baseline, it is important to note that depth is an intermediary that clearly eases the task by enabling the use of known geometric transformations.

\paragraph{Trade-off between adversarial and reconstruction loss:}
The training objective of the proposed refinement module involves the parameter $\lambda$ that trades-off the influence between the adversarial loss on simulated data and the reconstruction loss on either $\bevmapInit$ or OSM data.  In Figure~\ref{fig:bev_lambdasweep_examples} we demonstrate its impact qualitatively by setting $\lambda = \{0, 1, 5, 100, 500, 1000, 10^6\}$.  One can clearly see that higher $\lambda$ aligns better with actual image content.  However, setting $\lambda$ too high leads to replication of $\bevmapInit$ or over-fitting on $\bevmapOSM$.

\paragraph{Warping OSM data:}
In Table~\ref{tbl:bev_ablation_results_kittiinternal}, we compare different warping functions and optimization strategies for aligning the OSM data, as described in Section~\ref{sec:refine_bev}.  Our results show that the composition of ``Box'' (translation, scale and rotation) and ``Flow'' (displacement field) is superior to individual warps.  We can also see that the proposed alignment CNN trained jointly with the refinement module provides the best training signal from OSM data.  As already mentioned in Section~\ref{sec:refine_bev}, LBFG-S alignment failed for around 30\% of the training data, which explains the superiority of the proposed CNN for predicting warping parameters.

\paragraph{Qualitative results:}
Figure~\ref{fig:bev_qualitative_examples} demonstrates the beneficial impact of both the hallucination and refinement modules with several qualitative examples.  In the first three cases, we can observe the learned priors of the hallucination CNN that correctly handles largely occluded areas, which is evident from both the hallucinated semantics and the difference in the first two illustrated BEV maps (before and after hallucination).  Other examples illustrate how the refinement CNN completes unobserved areas and even completes whole side roads and intersections.  The last example shows a failure case of the hallucination step.  Videos for three sequences can be found at: \href{https://youtu.be/502Uen_Uv5M}{https://youtu.be/502Uen\_Uv5M} 

\begin{figure}[t]
\floatbox[{\capbeside\thisfloatsetup{capbesideposition={right,top},capbesidewidth=0.48\textwidth}}]{figure}[\FBwidth]
{\caption{Two examples of a BEV map including foreground objects, like cars here.  For each example, we also shows the input image, the semantic segmentation and the predicted depth map.}\label{fig:bev_dynamic_objects_example}}
{\includegraphics[width=0.48\textwidth]{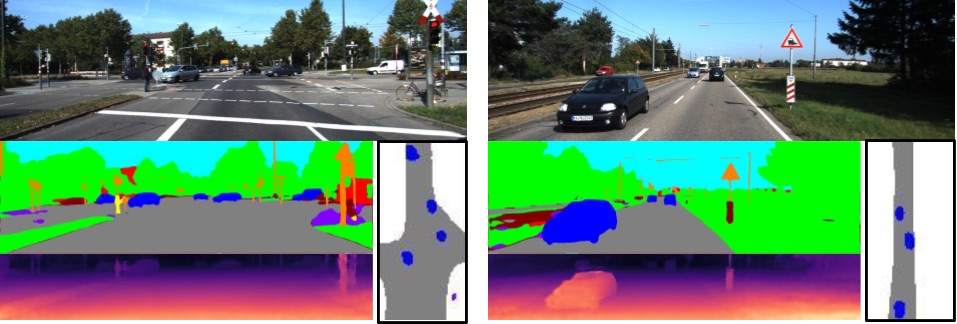}}
\vspace{-0.5cm}
\end{figure}

\subsection{Incorporating Foreground Objects into the BEV map}
\label{sec:exp_foreground_objects}
Finally, we show how foreground objects like cars or pedestrians can be handled in the proposed framework.  Since it is not the main focus of this paper, we use a simple baseline to lift 2D bounding boxes of cars into the BEV map.  Importantly, we demonstrate that our refinement module is able to handle foreground objects as well.  First we leverage the 3D ground truth annotations of the KITTI data set and estimate the mean dimensions of a 3D bounding box.  Then, for a given 2D bounding box in the perspective view, we use the estimated depth map to compute the 3D point of the bottom center of the bounding box, which is then used to translate our prior 3D bounding box in the BEV map.  The refinement network takes the initial BEV map that now includes foreground objects.  We extend the simulator to render objects as rectangles in the top-view and employ a self-reconstruction loss since OSM cannot provide such information.  Figure~\ref{fig:bev_dynamic_objects_example} gives two examples of the obtained BEV-map with foreground objects for illustrative purpose.
A full quantitative evaluation for localization accuracy and consistency with background requires significant extensions to be studied in our future work.


\section{Conclusion}
\label{sec:conclusion}

Our work addresses a complex problem in 3D scene understanding, namely, occlusion-reasoned semantic representation of outdoor scenes in the top-view, using just a single RGB image in the perspective view. This requires solving the canonical challenge of hallucinating semantics and geometry in areas occluded by foreground objects, for which we propose a CNN trained using only standard annotations in the perspective image. Further, we show that adversarial and warping-based refinement allow leveraging simulation and map data as valuable supervisory signals to learn prior knowledge. Quantitative and qualitative evaluations on the KITTI and Cityscapes datasets show attractive results compared to several baselines. While we have shown the feasibility of solving this problem using a single image, incorporating temporal information might be a promising extension for further gains. We finally note that with the use of indoor data sets like \cite{sam:Silberman12a,sam:Song15a}, along with simulators \cite{sam:Wu18a} and floor plans \cite{sam:Liu15a}, a similar framework may be derived for indoor scenes, which will be the subject of our future work.


\bibliographystyle{splncs}
\bibliography{myshortstrings,sam_schulter_2016_01_01}

\pagestyle{headings}
\mainmatter

\title{Supplemental Material:\\Learning to Look around Objects for Top-View Representations of Outdoor Scenes}

\titlerunning{Learning to Look around Objects for Top-View Representations of Road Scenes}
\authorrunning{Samuel Schulter, Menghua Zhai, Nathan Jacobs, Manmohan Chandraker}
\author{Samuel Schulter$^{1,\dagger}$ $\quad$ Menghua Zhai$^{2,\dagger}$ $\quad$ Nathan Jacobs$^2$\\ Manmohan Chandraker$^{1,3}$}
\institute{NEC-Labs$^1$, Computer Science University of Kentucky$^2$, UC San Diego$^3$}

\maketitle

The supplemental material contains the following items:
\begin{itemize}
\item \textbf{Section~\ref{sec:semseg_depth_bg}:} Additional experiments and details of the proposed hallucination models for semantic segmentation and depth prediction.
\item \textbf{Section~\ref{sec:refinement_ablation}:} Extended ablation study of the proposed refinement module.
\item \textbf{Section~\ref{sec:qualitative_results}:} Additional qualitative results of the semantic bird's eye view representation.
\item \textbf{Section~\ref{sec:semseg_depth_fg}:} Evaluation of semantic segmentation (and depth prediction) of foreground pixels.
\end{itemize}

\section{Semantic and depth hallucination}
\label{sec:semseg_depth_bg}
Hallucinating the semantics and depth behind foreground objects is an important part of the proposed BEV mapping.  The main paper shows two experiments for hallucination on the KITTI-Anon data set.  Here, we provide an ablation study of several other aspects of the hallucination-CNN on two data sets.

First, we investigate the random box sampling strategy used for training the hallucination networks, see Section 3.1 and Figure 2(b) in the main paper.  We name each sampling strategy based on four properties as ``geometry - background class - object size - object count'', where each property can take the following values:
\begin{itemize}
\item \textbf{``geometry''} is either ``none'' or ``perspective (persp.)''.  Perspective means that we apply a transformation to bounding boxes in order to mimic depth, \eg, objects further away become smaller.  ``None'' means that we do not change box sizes based on any prior.
\item \textbf{``background class''} is also a prior about where bounding boxes are placed.  For ``road'', we only put boxes at positions where they significantly overlap with road pixels.  For ``bg'', significant overlap with any background class is required.
\item \textbf{``object size''} is the typical object height closest to the camera, \ie, bottom of the image in the 2D image.  Note that object size is changed based on the y-axis location if the ``perspective'' option is used.
\item \textbf{``object count''} is the number of bounding boxes sampled per image.
\end{itemize}
In general, we can see from Table~\ref{tbl:semseg_depth_bg_masking} that placing more or bigger bounding boxes on images during the training process is beneficial, particularly for the hidden pixels which we are most interested in.  For instance, putting artificial boxes only on road pixels ``persp-road-150-3'', making the boxes small ``persp-bg-50-5'' or only sampling a single box ``persp-bg-100-1'' clearly deteriorates the performance of both semantic segmentation and depth prediction for hidden pixels.

Since the hallucination CNN is jointly trained for semantic segmentation and depth prediction, we also investigate the impact of joint training of the two related tasks.  We balance the loss functions for segmentation and depth prediction via $\lossSym = \lambda \cdot \lossSym_{\textrm{Depth}} + (1-\lambda) \cdot \lossSym_{\textrm{Seg}}$, where $\lambda \in [0,1]$.  We can see from Table~\ref{tbl:semseg_depth_bg_lambda} that the two tasks typically help each other, \ie, a value of $\lambda$ not at the two ends of the spectrum gives the best results.  The only exception is semantic segmentation of visible pixels on the Cityscapes data set.  For hidden pixels, the benefits of jointly training for both tasks can be seen on all data sets and is more pronounced than for visible pixels.

\begin{table}\centering\scriptsize
  \begin{tabular}{l l c c c c c c c c c c}
  \toprule
  Dataset         & Method                  & \multicolumn{8}{c|}{random-boxes}                          & \multicolumn{2}{c}{human-gt}  \\
                  &                         & \multicolumn{4}{c}{visible} & \multicolumn{4}{|c|}{hidden} & \multicolumn{1}{c|}{visible} & \multicolumn{1}{c}{hidden}  \\
                  &                         & iou   & RMSE   & acc   & \multicolumn{1}{c|}{ard}                  & iou & RMSE & acc & \multicolumn{1}{c|}{ard}                   & \multicolumn{1}{c|}{iou}     & iou \\
  \toprule
  \textbf{KITTI-} & none-bg-150-3           & \textbf{76.68} & \textbf{3.846} & \textbf{89.17} & \textbf{.0923}     & \textbf{64.63} & \textbf{5.360} & \textbf{74.43} & \textbf{.1413}                & \textbf{81.12} & 60.06 \\
  \textbf{Anon}   & persp-bg-150-3          & 75.37 & 4.510 & 88.39 & .0938     & 61.06 & 7.544 & 63.54 & .1748                & 80.21 & 60.19 \\
                  & persp-road-150-3        & 75.03 & 4.192 & 87.89 & .0964     & 49.18 & 7.848 & 61.09 & .1890                & 80.19 & 53.18 \\
                  & persp-bg-50-5           & 75.94 & 4.081 & 88.34 & .0943     & 53.01 & 8.375 & 57.83 & .1979                & 80.20 & 57.50 \\
                  & persp-bg-100-1          & 76.09 & 4.066 & 88.19 & .0943     & 57.27 & 8.658 & 58.17 & .1959                & 80.41 & 58.11 \\
                  & persp-bg-100-10         & 75.80 & 4.127 & 87.63 & .0963     & 59.22 & 8.177 & 60.72 & .1864                & 79.91 & \textbf{61.92} \\
  \midrule
  \textbf{KITTI-} & none-bg-150-3           & \textbf{70.95} & 2.411 & 91.65 & .0897     & 59.38 & 1.886 & 96.76 & 0.063                & \textbf{88.70} & \textbf{65.36} \\
  \textbf{Ros}    & persp-bg-150-3          & 70.82 & 2.298 & \textbf{92.44} & .0843     & \textbf{61.90} & 1.715 & 97.15 & 0.060                & 88.22 & 61.68 \\
                  & persp-road-150-3        & 69.14 & 2.394 & 91.65 & .0866     & 47.13 & 2.138 & 93.61 & 0.079                & 86.39 & 37.55 \\
                  & persp-bg-50-5           & 70.68 & 2.356 & 91.95 & .0858     & 51.92 & 2.389 & 92.23 & 0.079                & 88.08 & 53.66 \\
                  & persp-bg-100-1          & 70.78 & 2.295 & 92.32 & .0841     & 54.80 & 1.999 & 95.67 & 0.065                & 88.00 & 53.38 \\
                  & persp-bg-100-10         & 70.39 & \textbf{2.283} & 92.41 & \textbf{.0837}     & 61.64 & \textbf{1.698} & \textbf{97.26} & \textbf{0.056}                & 87.84 & 57.69 \\
  \bottomrule
\end{tabular}


  \caption{Impact of different masking strategies for training the proposed hallucination CNN for both semantic segmentation and depth prediction.}
  \label{tbl:semseg_depth_bg_masking}
\end{table}

\begin{table}\centering\scriptsize
  \begin{tabular}{l l c c c c c c c c c c}
  \toprule
  Dataset         & Method                  & \multicolumn{8}{c|}{random-boxes}                          & \multicolumn{2}{c}{human-gt}  \\
                  &                         & \multicolumn{4}{c}{visible} & \multicolumn{4}{|c|}{hidden} & \multicolumn{1}{c|}{visible} & \multicolumn{1}{c}{hidden}  \\
                  &                         & iou   & RMSE   & acc   & \multicolumn{1}{c|}{ard}                  & iou & RMSE & acc & \multicolumn{1}{c|}{ard}                   & \multicolumn{1}{c|}{iou}     & iou \\
  \toprule
  \textbf{KITTI-} & 0.00                    & 72.92 &     - &     - &     -     & 62.48 &     - &     - &     -                & 79.45 & 59.78 \\
  \textbf{Anon}   & 0.25                    & 75.17 & 4.105 & 87.64 & .1000     & 63.70 & 5.905 & 70.79 & .1577                & 80.24 & \textbf{60.26} \\
                  & 0.50                    & \textbf{76.20} & 3.832 & 89.44 & .0909     & \textbf{64.41} & 5.503 & 74.31 & .1446                & \textbf{80.85} & 59.39 \\
                  & 0.75                    & 75.29 & \textbf{3.778} & \textbf{90.17} & \textbf{.0873}     & 63.33 & \textbf{5.334} & \textbf{76.02} & \textbf{.1380}                & 80.72 & 59.94 \\
                  & 1.00                    &     - & 3.921 & 88.79 & .0954     &     - & 5.428 & 74.73 & .1452                &     - &     - \\
  \midrule
  \textbf{KITTI-} & 0.00                    & 69.29 & -     & -     & -         & 57.25 & -     & -     & -                    & 87.93 & 53.29 \\
  \textbf{Ros}    & 0.25                    & 70.05 & 2.496 & 91.05 & .0918     & \textbf{60.95} & 1.940 & 95.78 & .0670                & 87.59 & \textbf{56.92} \\
                  & 0.50                    & 71.26 & 2.278 & 92.84 & .0816     & 59.54 & 1.906 & 96.58 & .0606                & 88.05 & 54.90 \\
                  & 0.75                    & \textbf{71.39} & \textbf{2.179} & \textbf{93.58} & \textbf{.0769}     & 59.89 & \textbf{1.729} & \textbf{97.24} & \textbf{.0570}                & \textbf{88.71} & 53.77 \\
                  & 1.00                    & -     & 2.649 & 91.16 & .0896     & -     & 2.166 & 94.65 & .0732                & - & - \\
  \midrule
  \textbf{City-}  & 0.00                    & \textbf{71.14} & -      & -     & -         & 59.80 & -     & -     & -                    & \textbf{74.00} & 60.27 \\
  \textbf{scapes} & 0.25                    & 70.62 & 12.810 & 84.57 & .1325     & \textbf{60.26} & 8.290 & 85.39 & .123                 & 73.72 & 60.57 \\
                  & 0.50                    & 69.75 & 12.813 & 86.27 & .1234     & 58.80 & 8.226 & 86.83 & .116                 & 73.49 & \textbf{61.24} \\
                  & 0.75                    & 68.19 & \textbf{12.787} & \textbf{87.03} & \textbf{.1206}     & 55.69 & 8.187 & \textbf{87.24} & \textbf{.112}                 & 72.85 & 60.89 \\
                  & 1.00                    & -     & 12.796 & 84.63 & .1333     & -     & \textbf{7.848} & 85.18 & .127                 & -     & -     \\
  \bottomrule
\end{tabular}


  \caption{The impact of the trade-off between semantics and depth when training the hallucination CNN jointly for both tasks.}
  \label{tbl:semseg_depth_bg_lambda}
\end{table}

\begin{figure}\centering
  \includegraphics[width=0.48\textwidth]{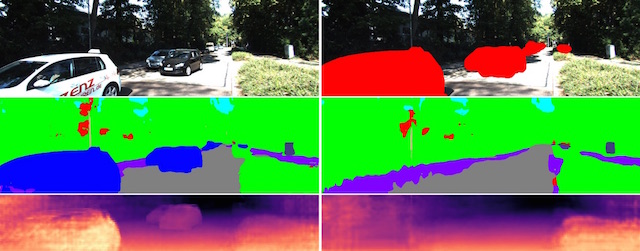}
  \includegraphics[width=0.48\textwidth]{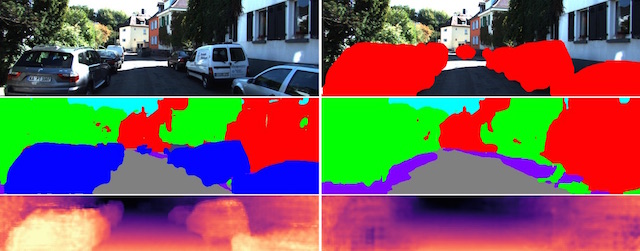}\\
  \quad\\
  \includegraphics[width=0.48\textwidth]{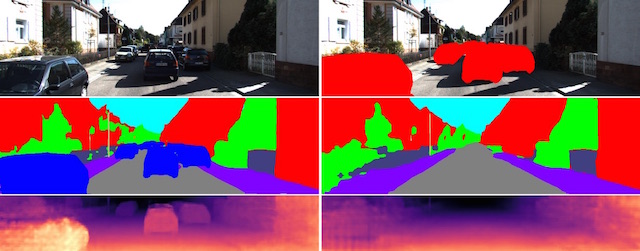}
  \includegraphics[width=0.48\textwidth]{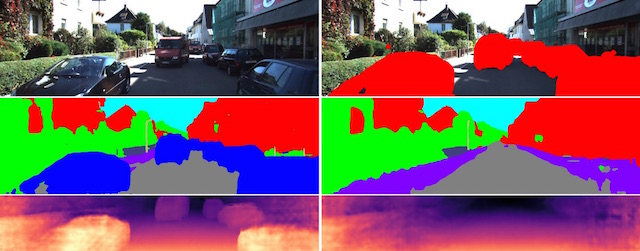}\\
  \quad\\
  \includegraphics[width=0.48\textwidth]{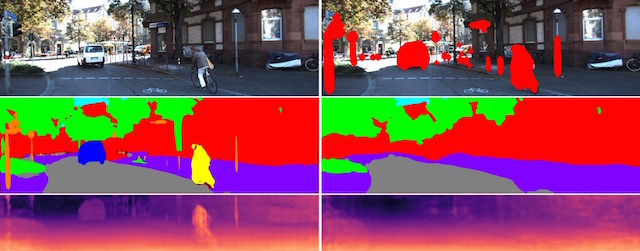}
  \includegraphics[width=0.48\textwidth]{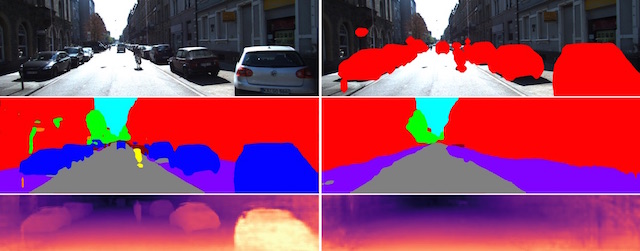}\\
  \quad\\
  \includegraphics[width=0.48\textwidth]{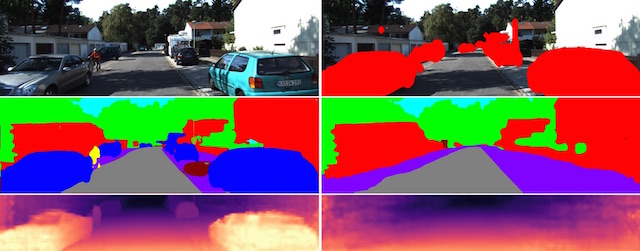}
  \includegraphics[width=0.48\textwidth]{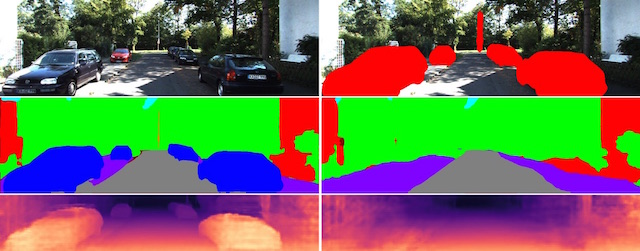}\\
  \quad\\
  \includegraphics[width=0.48\textwidth]{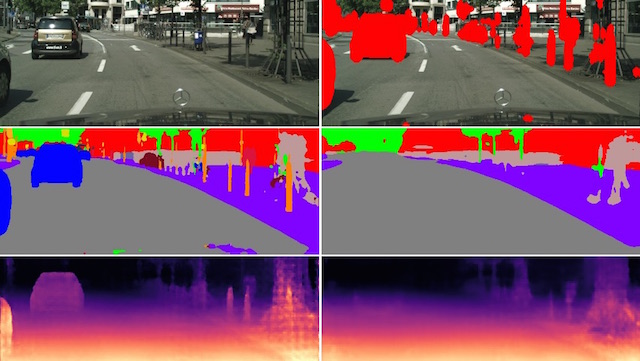}
  \includegraphics[width=0.48\textwidth]{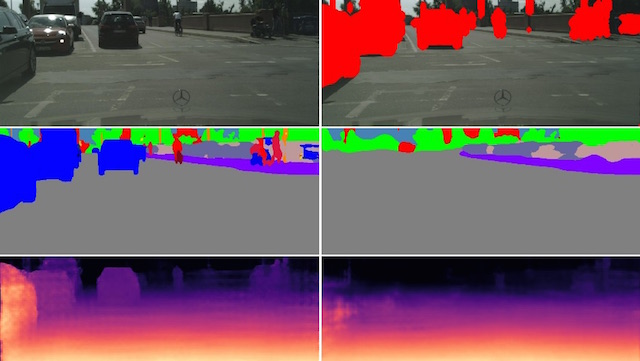}
  \caption{Ten qualitative results of the hallucination CNN (two per row).  The first column in each example shows the input RGB image, the semantic segmentation and depth prediction of all visible pixels, \ie, including foreground objects.  The second column shows the masked RGB image and the hallucinated semantics and depth.  Note how foreground objects like cars or pedestrians are replaced with learned priors about the environment.  The first two rows are examples from the KITTI-Ros data set.  Rows 3 and 4 are from the KITTI-Anon data set and the last row is from Cityscapes.}
  \label{fig:qualit_results_hallucination}
\end{figure}

In Figure~\ref{fig:qualit_results_hallucination} we present additional qualitative results of hallucinating semantics and depth, which are contrasted with the corresponding standard foreground semantic segmentation and depth prediction.  One can clearly see the prior knowledge learned by the hallucination CNN.

\section{Additional results of the refinement module}
\label{sec:refinement_ablation}
\begin{figure}[h]\centering
  \includegraphics[width=1.0\textwidth]{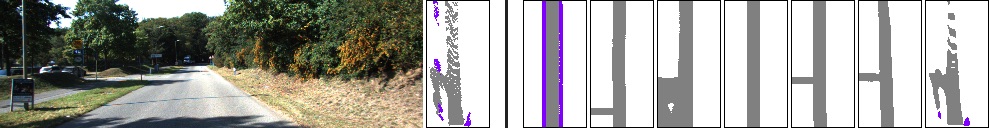}
  \includegraphics[width=1.0\textwidth]{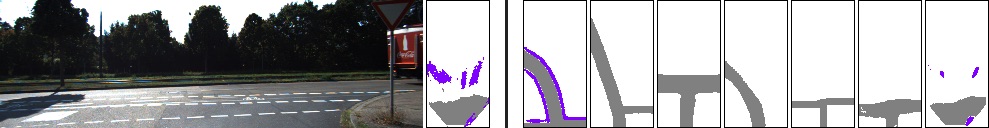}
  \includegraphics[width=1.0\textwidth]{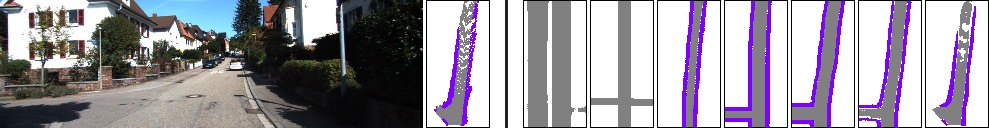}
  \includegraphics[width=1.0\textwidth]{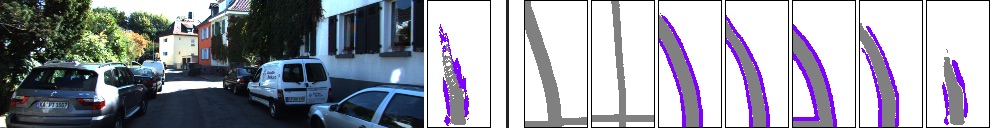}
  \includegraphics[width=1.0\textwidth]{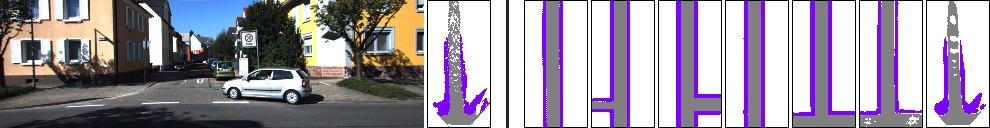}
  \caption{Five examples of the influence of the trade-off between adversarial and reconstruction loss in the proposed refinement module.  Each example shows the input RGB image and the initial BEV map (\ie, after hallucinating depth and semantics and mapping into the top-view) on the left.  The seven BEV-maps on the right show the output of the refinement module trained with varying $\lambda = \{0,1,5,100,500,1000,16^6\}$ (from left to right).  One can clearly observe that setting $\lambda$ too low or too high results in unfavorable representations.}
  \label{fig:qualitative_results_lambda_refinement}
\end{figure}

In this section, we show additional results of the impact of the trade-off between the adversarial loss (on simulated data) and the reconstruction loss (with the initial BEV map or OSM data if available).  The main paper contains one example (in Figure 7).  Here, we provide more examples in Figure~\ref{fig:qualitative_results_lambda_refinement}.  It is clearly evident from the figure that the reconstruction loss needs to be properly balanced with the adversarial loss.  No reconstruction loss obviously leads to generating scene layouts that don't match the actual image evidence.  On the other hand, putting too much weight on the reconstruction loss leads the refinement module to learning the identity function without improving upon its input.

\section{Additional qualitative results}
\label{sec:qualitative_results}
\begin{figure}[t]\centering
  \includegraphics[width=0.24\textwidth]{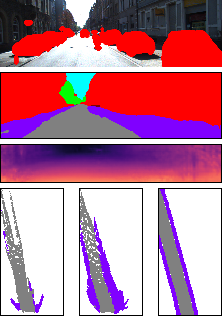}
  \includegraphics[width=0.24\textwidth]{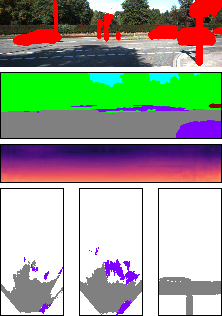}
  \includegraphics[width=0.24\textwidth]{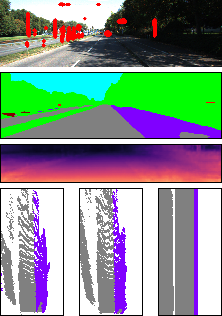}
  \includegraphics[width=0.24\textwidth]{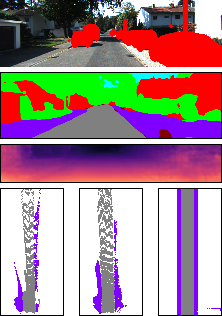}\\
  \quad\\
  \includegraphics[width=0.24\textwidth]{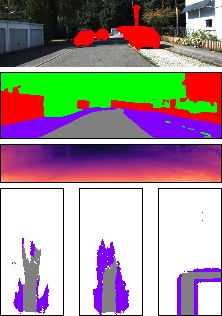}
  \includegraphics[width=0.24\textwidth]{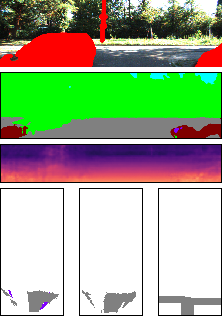}
  \includegraphics[width=0.24\textwidth]{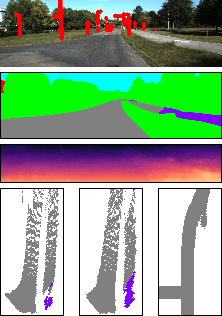}
  \includegraphics[width=0.24\textwidth]{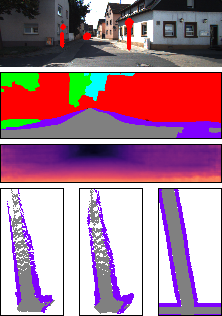}\\
  \caption{Qualitative results of the final semantic BEV representation for the KITTI-Anon data set (four examples per row).  Each example shows the masked RGB input, the hallucinated semantics and depth, as well as three BEV maps, which are (from left to right): The BEV map without hallucination, with hallucination, and after refinement.}
  \label{fig:qualit_examples_kitti_internal}
\end{figure}

\begin{figure}[t]\centering
  \includegraphics[width=0.24\textwidth]{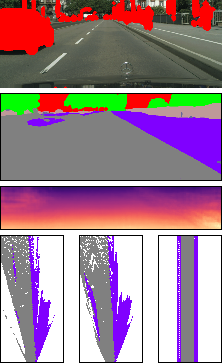}
  \includegraphics[width=0.24\textwidth]{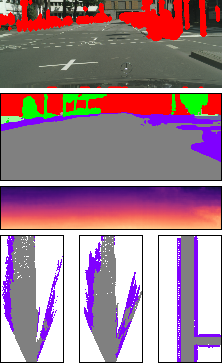}
  \includegraphics[width=0.24\textwidth]{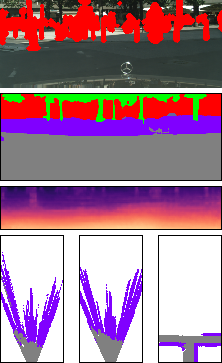}
  \includegraphics[width=0.24\textwidth]{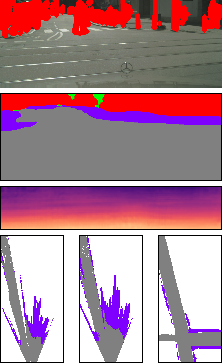}\\
  \quad\\
  \includegraphics[width=0.24\textwidth]{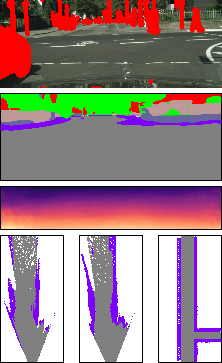}
  \includegraphics[width=0.24\textwidth]{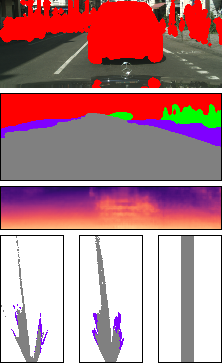}
  \includegraphics[width=0.24\textwidth]{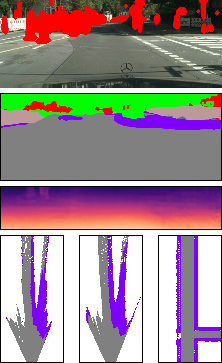}
  \includegraphics[width=0.24\textwidth]{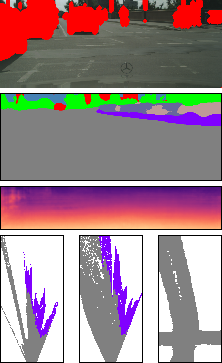}
  \caption{Qualitative results of the final semantic BEV representation for the Cityscapes data set (four examples per row).  Each example shows the masked RGB input, the hallucinated semantics and depth, as well as three BEV maps, which are (from left to right): The BEV map without hallucination, with hallucination, and after refinement.}
  \label{fig:qualit_examples_cityscapes}
\end{figure}

The main paper already contains qualitative results of the final semantic BEV representation for the KITTI-Ros data set in Figure 8.  Here, we provide additional examples for the KITTI-Anon and the Cityscapes data sets in Figures~\ref{fig:qualit_examples_kitti_internal} and \ref{fig:qualit_examples_cityscapes}, respectively.

\begin{figure}
  \centering
 \includegraphics[width=1.0\linewidth]{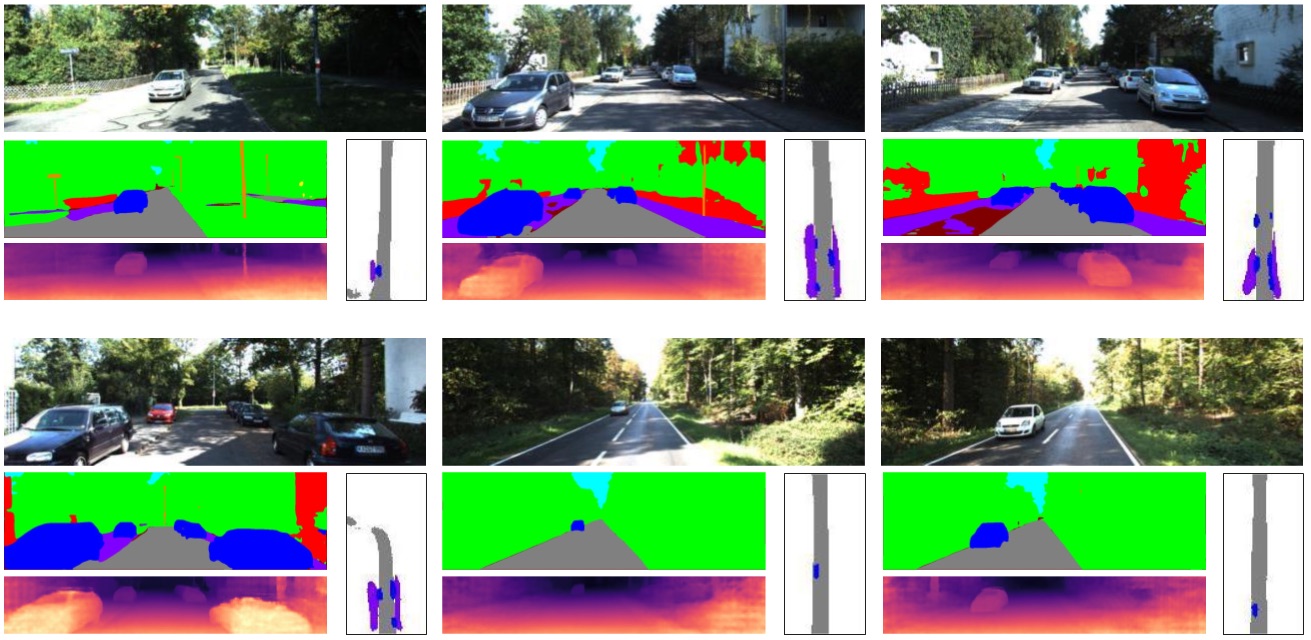}
  \caption{Qualitative examples including foreground objects.  Each example shows the RGB input image, the semantic segmentation, the depth map, and the estimated BEV representation.}
  \label{fig:qualit_examples_dynamic_objects}
\end{figure}

Moreover, we show a few additional examples of our representation including dynamic foreground objects like cars in Figure~\ref{fig:qualit_examples_dynamic_objects}.


\section{Semantic segmentation and depth prediction of visible pixels}
\label{sec:semseg_depth_fg}
The proposed semantic bird's eye view representation assumes a semantic segmentation of the visible pixels as input in order to identify foreground objects which define occlusions of the scene.  Any semantic segmentation module can be used and we picked a CNN architecture inspired by the PSP module~\cite{sam:Zhao17a}.  Besides standard semantic segmentation, this CNN also predicts depth of all visible pixels with a second decoder similar in structure to~\cite{sam:Laina16a}, which is required to estimate depth for 3D localization of dynamic foreground objects (or traffic participants) like cars and pedestrians.

\begin{figure}[h]
  \centering
  \includegraphics[width=1.0\linewidth]{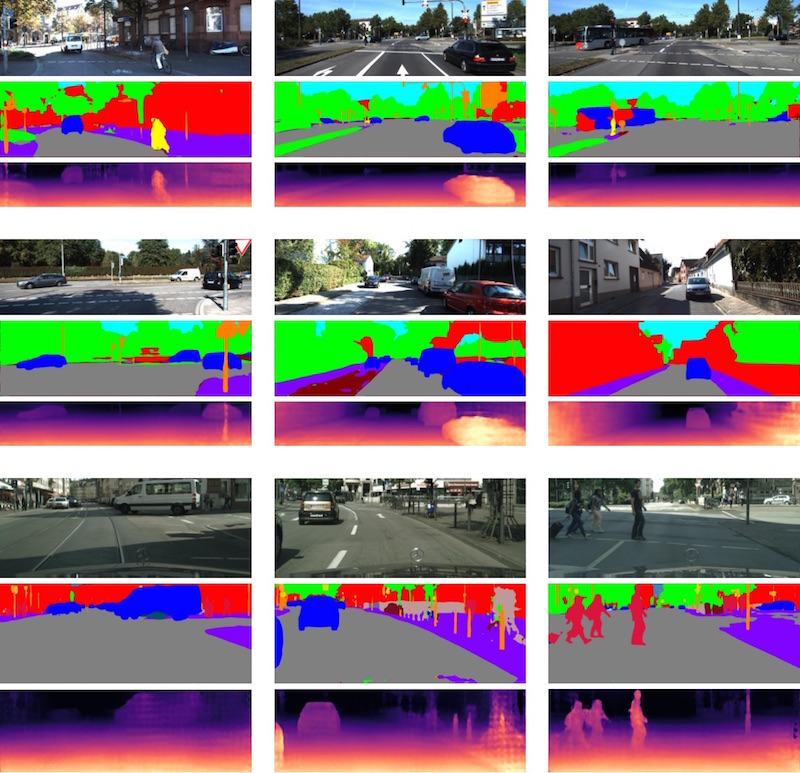}
  \caption{Qualitative examples for semantic segmentation and depth prediction. First two rows: KITTI; Last row: Cityscapes.}
  \label{fig:semseg_depth_fg}
\end{figure}

For completeness, this section provides a quantitative evaluation of this CNN.  Table~\ref{tbl:semseg_depth_fg} shows our results for semantic segmentation and depth prediction and Figure~\ref{fig:semseg_depth_fg} provides some qualitative examples.  For evaluating semantic segmentation we use mean IoU as in the main paper.  For evaluating depth prediction, we present additional metrics that are typically used and defined in~\cite{sam:Eigen14a}: RMSE, RMSE-log, accuracy (with threshold of $\delta = 1.25$), and absolute relative difference (ARD).  Note that we use a down-scaled version of Cityscapes (by a factor of $0.625$) for this experiment because it significantly decreases runtime and memory consumption during training and evaluation.

\begin{table}\centering\footnotesize
  \begin{tabular}{l c c c c c}
  \toprule
  Dataset                                    & mIoU     & RMSE  & RMSE-log  & ACC   & ARD      \\
  \midrule
  \textbf{KITTI-Anon}                        & 69.63    & 4.129 & 0.158     & 89.56 & .0928    \\
  \textbf{KITTI-Ros}~\cite{sam:Ros15a}       & 59.02    & 3.976 & 0.172     & 86.73 & .1076    \\
  \textbf{Cityscapes}~\cite{sam:Cordts16a}   & 63.66    & 8.352 & 0.220     & 89.65 & .1113    \\
  \bottomrule
\end{tabular}

  \caption{Quantitative evaluation of standard semantic segmentation (mIoU) and depth prediction (RMSE, RMSE-log, ACC, ARD) for all visible pixels, \ie, the first step in the proposed pipeline for BEV-mapping. We show results for all three data sets used in the main paper.}
  \label{tbl:semseg_depth_fg}
\end{table}

\bibliographystyle{splncs}
\bibliography{myshortstrings,sam_schulter_2016_01_01}

\end{document}